\documentclass[fleqn,10pt]{wlscirep}
\pdfoutput=1
\usepackage{amsmath}
\usepackage{amssymb}
\usepackage{bm}
\usepackage[font={sf, sl, small}, labelfont={sf,bf}, format=plain,justification=justified
]{caption}

\usepackage[english]{babel}
\usepackage[margins]{trackchanges}
\addeditor{GAB}
\addeditor{DOR}

\title{Deep learning cardiac motion analysis for human survival prediction}

\author[1,+]{Ghalib A. Bello}
\author[1,2,+]{Timothy J.W. Dawes}
\author[1,3]{Jinming Duan}
\author[1,3]{Carlo Biffi}
\author[1]{Antonio de Marvao}
\author[4] {Luke S. G. E. Howard}
\author [2,4] {J. Simon R. Gibbs}
\author [6] {Martin R. Wilkins}
\author [1,2,5] {Stuart A. Cook}
\author[3]{Daniel Rueckert}
\author[1,*]{Declan P. O'Regan}
\affil[1]{MRC London Institute of Medical Sciences, Imperial College London,UK}
\affil[2]{National Heart and Lung Institute, Imperial College London, UK}
\affil[3]{Department of Computing, Imperial College London, UK}
\affil [4] {Imperial College Healthcare NHS Trust, London, UK} 
\affil [5] {National Heart Centre Singapore, Singapore, and Duke-NUS Graduate Medical School, Singapore}
\affil [6] {Division of Experimental Medicine, Department of Medicine, Imperial College London, UK}
\affil[*]{declan.oregan@imperial.ac.uk}

\affil[+]{These authors contributed equally to this work}

\begin{abstract}
Motion analysis is used in computer vision to understand the behaviour of moving objects in sequences of images. Optimising the interpretation of dynamic biological systems  requires accurate and precise motion tracking as well as efficient representations of high-dimensional motion trajectories so that these can be used for prediction tasks. Here we use image sequences of the heart, acquired using cardiac magnetic resonance imaging, to create time-resolved three-dimensional segmentations using a fully convolutional network trained on anatomical shape priors. This dense motion model formed the input to a supervised denoising autoencoder (4D\textit{survival}), which is a hybrid network consisting of an autoencoder that learns a task-specific latent code representation trained on observed outcome data, yielding a latent representation optimised for survival prediction. To handle right-censored survival  outcomes, our network used a Cox partial likelihood loss function. In a study of 302 patients the predictive accuracy (quantified by Harrell's C-index) was significantly higher (p $< .0001$) for our model C=0.73 (95$\%$ CI: 0.68 - 0.78) than the human benchmark of C=0.59 (95$\%$ CI: 0.53 - 0.65). This work demonstrates how a complex computer vision task using high-dimensional medical image data can efficiently predict human survival.\\


\end{abstract}

\begin{document}

\flushbottom
\maketitle
\thispagestyle{empty}

\section*{Introduction}
Techniques for vision-based motion analysis aim to understand the behaviour of moving objects in image sequences.\cite{Wang2010} In this domain deep learning architectures have achieved a wide range of competencies for object tracking, action recognition, and semantic segmentation.\cite{deep-learning-intelligent-video-analysis} Making predictions about future events from the current state of a moving three dimensional (3D) scene depends on learning correspondences between patterns of motion and subsequent outcomes. Such relationships are important in biological systems which exhibit complex spatio-temporal behaviour in response to stimuli or as a consequence of disease processes. Here we use recent advances in machine learning for visual processing tasks to develop a generalisable approach for modelling time-to-event outcomes from time-resolved 3D sensory input. We tested this on the challenging task of predicting survival due to heart disease through analysis of cardiac imaging.  

The motion dynamics of the beating heart are a complex rhythmic pattern of non-linear trajectories regulated by molecular, electrical and biophysical processes.\cite{Liang2017} Heart failure is a disturbance of this coordinated activity characterised by adaptations in cardiac geometry and motion that lead to impaired organ perfusion.\cite{Savarese2017} For this prediction task we studied patients diagnosed with pulmonary hypertension (PH), characterised by right ventricular (RV) dysfunction, as this is a disease with high mortality where the choice of treatment depends on individual risk stratification.\cite{Galie2016} Our input data were derived from cardiac magnetic resonance (CMR) which acquires imaging of the heart in any anatomical plane for dynamic assessment of function. While explicit measurements of performance obtained from myocardial motion tracking detect early contractile dysfunction and act as discriminators of different pathologies,\cite{Puyol,Scatteia2017} we hypothesized that learned features of complex 3D cardiac motion would provide enhanced prognostic accuracy. 

A major challenge for medical image analysis has been to automatically derive quantitative and clinically-relevant information in patients with disease phenotypes. Our method employs a fully convolutional network (FCN) to learn a cardiac segmentation task from manually-labelled priors. The outputs are smooth 3D renderings of frame-wise cardiac motion which are used as input data to a supervised denoising autoencoder prediction network which we refer to as 4D\textit{survival}. The aim is to learn latent representations robust to noise and salient for survival prediction. We then compared our model to a benchmark of conventional human-derived volumetric indices in survival prediction.

\section*{Results}
\subsection*{Baseline Characteristics}
Data from all 302 patients with incident PH were included for analysis. Objective diagnosis was made according to haemodynamic criteria.\cite{Galie2016} Patients were investigated between 2004 and 2017, and were followed-up until November 27, 2017 (median 371 days). All-cause mortality was 28$\%$ (85 of 302). Table \ref{tab:sampchar} summarizes characteristics of the study sample at the date of diagnosis. No subjects' data were excluded. 

\begin{table}[!b]
\scriptsize
\renewcommand{\arraystretch}{1.3}
\centering
\begin{tabular}{p{6.5cm} r c}
\textbf{Characteristic} & \textbf{n} & \textbf{$\%$ or Mean $\pm$SD} \\
\hline

Age (years) & & 62.9 $\pm$14.5 \\
Body surface area ($m^2$) & & 1.92 $\pm$0.25 \\
Male & 169 & 56 \\
Race & &  \\
\hspace{3mm}Caucasian & 215 & 71.2 \\
\hspace{3mm}Asian & 7 & 2.3 \\
\hspace{3mm}Black & 13 & 4.3 \\
\hspace{3mm}Other & 28 & 9.3 \\
\hspace{3mm}Unknown & 39 & 12.9 \\
WHO functional class & &  \\
\hspace{3mm}I   &   1 & 0 \\
\hspace{3mm}II  &  45 & 15 \\
\hspace{3mm}III & 214 & 71\\
\hspace{3mm}IV  &  42 & 14 \\
Haemodynamics & &  \\
\hspace{3mm}Systolic BP (mmHg)   &   & 131.5 $\pm$25.2 \\
\hspace{3mm}Diastolic BP (mmHg)   &   & 75 $\pm$13 \\
\hspace{3mm}Heart rate (beats/min)   &  & 69.8 $\pm$ 22.5  \\
\hspace{3mm}Mean right atrial pressure (mmHg)   &   & 9.9 $\pm$5.8 \\
\hspace{3mm}Mean pulmonary artery pressure (mmHg)  &  & 44.1 $\pm$12.6 \\
\hspace{3mm}Pulmonary vascular resistance (Wood units) & & 8.9 $\pm$5.0 \\
\hspace{3mm}Cardiac output (l/min) &  & 4.3 $\pm$1.5 \\
LV Volumetry & &  \\
\hspace{3mm}LV ejection fraction (\%)   & & 61 $\pm$ 11.1 \\
\hspace{3mm}LV end diastolic volume index (ml/m)  & & 110 $\pm$ 37.4 \\
\hspace{3mm}LV end systolic volume index (ml/m) & & 44 $\pm$ 22.9 \\
RV Volumetry & &  \\
\hspace{3mm}RV ejection fraction (\%)   &  & 38 $\pm$ 13.7 \\
\hspace{3mm}RV end diastolic volume (ml/m)  &  & 194 $\pm$ 62 \\
\hspace{3mm}RV end systolic volume (ml/m) & & 125 $\pm$ 59.3 \\
\hline
\end{tabular}

\caption{\label{tab:sampchar}Patient characteristics at baseline (date of MRI scan). WHO, World Health Organization; BP, Blood pressure; LV, left ventricle; RV, right ventricle.}
\end{table}

\subsection*{MR Image Processing}
\begin{figure}[!ht]
\centering
\includegraphics[width=\linewidth]{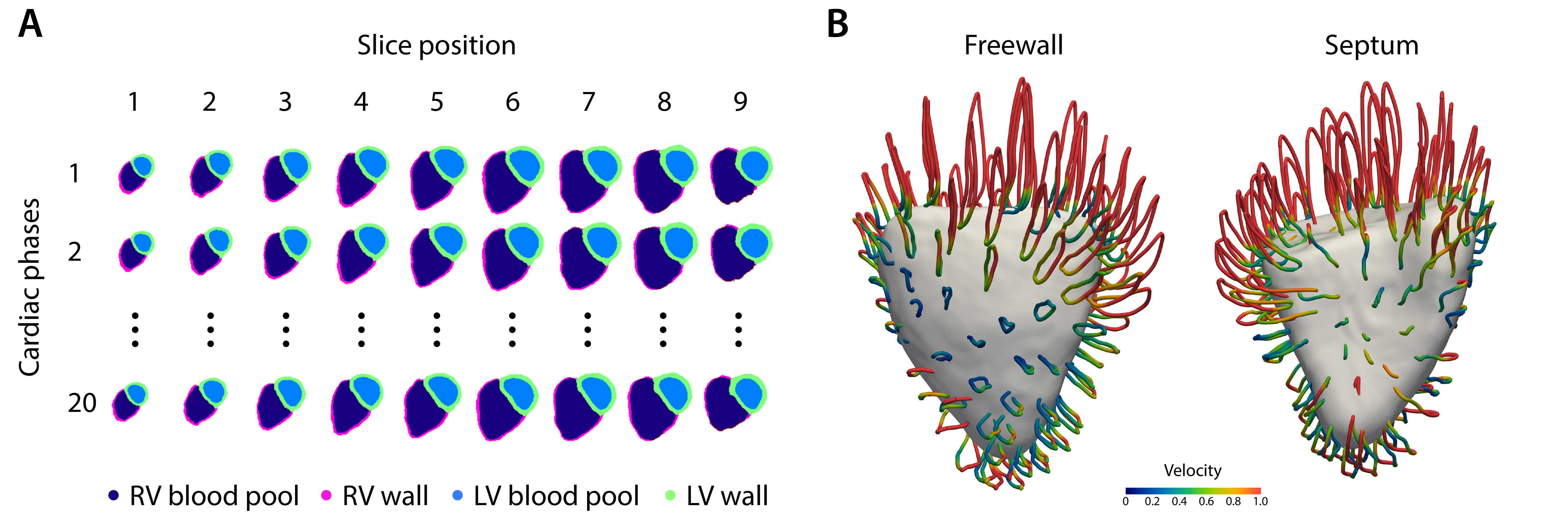}
\caption{A) An example of an automatic cardiac image segmentation of each short-axis cine image from apex (slice 1) to base (slice 9) across 20 temporal phases. Data were aligned to a common reference space to build a population model of cardiac motion. B) Trajectory of right ventricular contraction and relaxation averaged across the study population plotted as looped pathlines for a sub-sample of 100 points on the heart (magnification factor of x4). Colour represents relative myocardial velocity at each phase of the cardiac cycle. A surface-shaded model of the heart is shown at end-systole. These dense myocardial motion fields for each patient were used as an input to the prediction network. LV, left ventricular; RV, right ventricular.}
\label{fig:surfacemesh_mag}
\end{figure}

Automatic segmentation of the ventricles from gated CMR images was performed for each slice position at each of 20 temporal phases producing a total of 69,820 label maps for the cohort (Figure \ref{fig:surfacemesh_mag}A). Image registration was used to track the motion of corresponding anatomic points. Data for each subject was aligned producing a dense model of cardiac motion across the patient population (Figure \ref{fig:surfacemesh_mag}B) which was then used as an input to the 4D\textit{survival} network.

\subsection*{Predictive performance}
\begin{figure}[!ht]
\centering
\includegraphics[width=\linewidth]{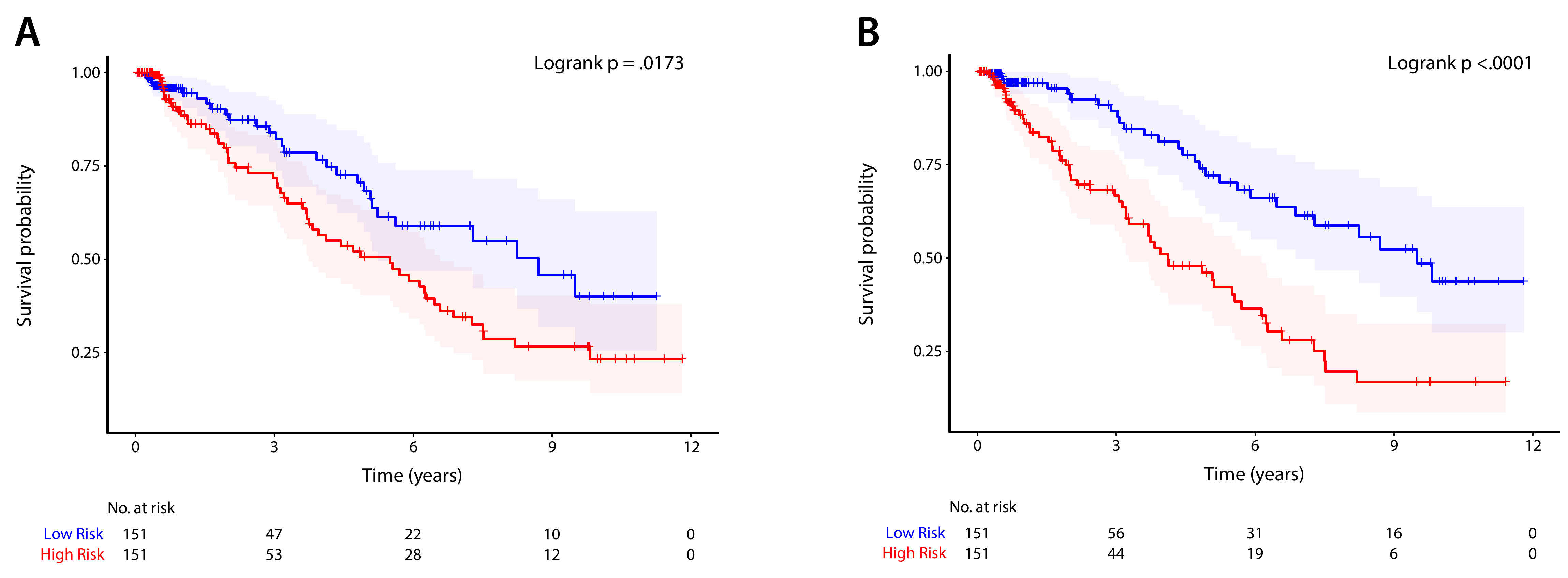}
\caption{Kaplan-Meier plots for A) a conventional parameter model using a composite of manually-derived volumetric measures, and B) a deep learning prediction model (4Dsurvival) whose input was time-resolved three dimensional models of cardiac motion. \\For both models, patients were divided into low- and high-risk groups by median risk score. Survival function estimates for each group (with $95\%$ confidence intervals) are shown. For each plot, the Logrank test was performed to compare survival curves between risk groups (conventional parameter model: ${\chi}^{2}=5.7, p=.0173$; \textit{4Dsurvival}: ${\chi}^{2}=20.7, p<.0001$)}
\label{fig:KMplots}
\end{figure}

Bootstrapped internal validation was applied to the 4D\textit{survival} and benchmark conventional parameter models. The apparent predictive accuracy for 4D\textit{survival} was $C=0.85$ and the optimism-corrected value was $C=$ 0.73 ($95\%$ CI: 0.68-0.78). For the benchmark conventional parameter model, the apparent predictive accuracy was $C=0.61$ with the corresponding optimism-adjusted value being $C=$ 0.59 ($95\%$ CI: 0.53-0.65). The accuracy for 4D\textit{survival} was significantly higher than that of the conventional parameter model (p $< .0001$). After bootstrap validation, a final model was created using the training and optimization procedure outlined in the \textit{Methods} section with Kaplan-Meier plots showing the survival probability estimates over time, stratified by risk groups defined by each model's predictions (Figure \ref{fig:KMplots}). 

\subsection*{Visualization of Learned Representations}
\begin{figure}[ht]
\centering
\includegraphics[width=\linewidth]{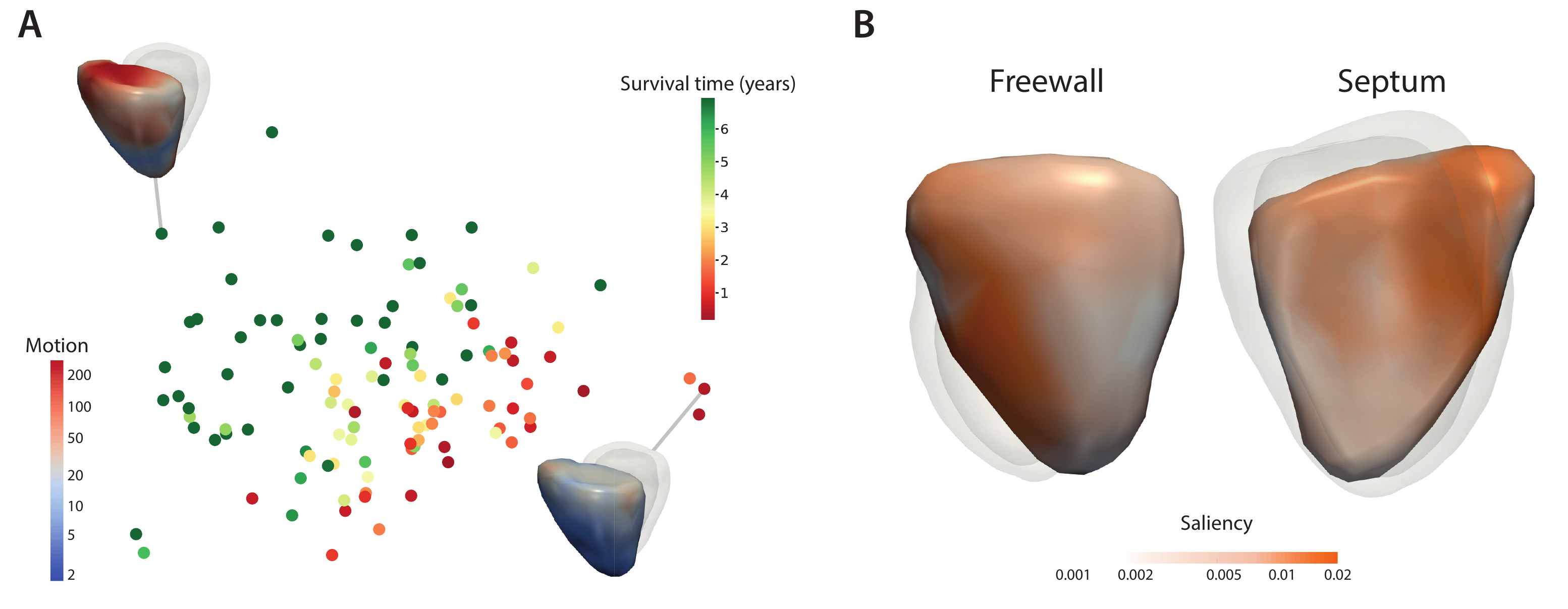}
\caption{A) A 2-dimensional projection of latent representations of cardiac motion in the 4D\textit{survival} network labelled by survival time. A visualisation of RV motion is shown for two patients with contrasting risks. B) Saliency map showing regional contributions to survival prediction by right ventricular motion. Absolute regression coefficients are expressed on a log-scale.}
\label{fig:pca1yr}
\end{figure}

To assess the ability of the 4D\textit{survival} network to learn discriminative features from the data, we examined the encoded representations by projection to 2D space using Laplacian Eigenmaps \cite{Belkin2002} (Figure \ref{fig:pca1yr}A). In this figure, each subject is represented by a point, the colour of which is based on the subject's survival time, i.e. time elapsed from baseline (date of MRI scan) to death (for uncensored patients), or to the most recent follow-up date (for censored patients). Survival time was truncated at 7 years for ease of visualization. As is evident from the plot, our network's compressed representations of 3D motion input data show distinct patterns of clustering according to survival time. Figure \ref{fig:pca1yr}A also shows visualizations of RV motion for 2 exemplar subjects at opposite ends of the risk spectrum.
We also assessed the extent to which motion in various regions of the RV contributed to overall survival prediction. Fitting univariate linear models to each vertex in the mesh (see Methods for full details), we computed the association between the magnitude of cardiac motion and the 4D\textit{survival} network's predicted risk score, yielding a set of regression coefficients (one per vertex) that were then mapped onto a template RV mesh, producing a 3D saliency map (Figure \ref{fig:pca1yr}B). These show the contribution from spatially distant but functionally synergistic regions of the RV in influencing survival in PH. 

\section*{Discussion}

Machine learning algorithms have been used in a variety of motion analysis tasks from classifying complex traits to predicting future events from a given scene.\cite{Li2017,Walker2016,Butepage2017} We show that compressed representations of a dynamic biological system moving in 3D space offer a powerful approach for time-to-event analysis. In this example we demonstrate the effectiveness of a deep learning algorithm, trained to find correspondences between heart motion and patient outcomes, for efficiently predicting human survival.  

The traditional paradigm of epidemiological research is to draw insight from large-scale clinical studies through linear regression modelling of conventional explanatory variables, but this approach does not embrace the dynamic physiological complexity of heart disease.\cite{Johnson2017} Even objective quantification of heart function by conventional analysis of cardiac imaging relies on crude measures of global contraction that are only moderately reproducible and insensitive to the underlying disturbances of cardiovascular physiology.\cite{Cikes2016} Integrative approaches to risk classification have used unsupervised clustering of broad clinical variables to identify heart failure patients with distinct risk profiles, \cite{Ahmad2014,Shah2015} while supervised machine learning algorithms can diagnose, risk stratify and predict adverse events from health record data.\cite{Awan2018,Tripoliti2017} In the wider health domain deep learning has achieved successes in forecasting survival from high-dimensional inputs such as cancer genomic profiles and gene expression data,\cite{Yousefi2017,Ching2018b} and in formulating personalised treatment recommendations.\cite{Katzman2018}

With the exception of natural image tasks, such as classification of skin lesions,\cite{Esteva2017} biomedical imaging poses a number of challenges for machine learning as the datasets are often of limited scale, inconsistently annotated, and typically high-dimensional.\cite{Ching2018} Architectures predominantly based on convolutional neural nets (CNNs), often using data augmentation strategies, have been successfully applied in computer vision tasks to enhance clinical images, segment organs and classify lesions.\cite{Litjens2017,Shen2017} Segmentation of cardiac images in the time domain is a well-established visual correspondence task that has recently achieved expert-level performance with FCN architectures.\cite{bai2018automated} Components of these cardiac motion models have prognostic utility,\cite{Dawes2017} and in this work we harness the power of deep learning to predict outcomes from dense phenotypic data in time-resolved segmentations of the heart.   

Autoencoding is a dimensionality reduction technique in which an encoder takes an input and maps it to a latent representation (lower-dimensional space) which is in turn mapped back to the space of the original input. The latter step represents an attempt to `reconstruct' the input from the compressed (latent) representation, and this is done in such a way as to minimise the reconstruction error, i.e. the degree of discrepancy between the input and its reconstructed version. Our algorithm is based on a denoising autoencoder (DAE), a type of autoencoder which aims to extract more robust latent representations by corrupting the input with stochastic noise.\cite{Rifai2011} While conventional autoencoders are used for unsupervised learning tasks we extend recent proposals for supervised autoencoders in which the learned representations are both reconstructive \textit{and} discriminative.\cite{Rolfe2013, Huang2016, Du2018, Zaghbani2018,BEAULIEUJONES2016,Shakeri2016,Biffi2018} We achieved this by adding a prediction branch to the network with a loss function for survival inspired by the Cox proportional hazards model. A hybrid loss function, optimising the trade-off between survival prediction and accurate input reconstruction, is calibrated during training. The compressed representations of 3D motion predict survival more accurately than a composite measure of conventional manually-derived parameters measured on the same images.  To safeguard against overfitting on our patient data we used dropout and $L_1$ regularization to yield a robust prediction model.

Our approach enables fully automated and interpretable predictions of survival from moving clinical images - a task that has not been previously achieved in heart failure or other disease domains. This fast and scalable method is readily deployable and could have a substantial impact on clinical decision making and understanding of disease mechanisms.
Further enhancement in predictive performance may be achievable by modelling multiple observations over time, for instance using long short-term memory (LSTM) and other recurrent neural network architectures,\cite{Bao2017,lim2018} and handling independent competing risks.\cite{Lee2018DeepHitAD} The next step towards implementation of these architectures is to train them on larger and more diverse multicentre patient groups using image data and other prognostic variables, before performing external validation of survival prediction in a clinical setting against a benchmark of established risk prediction scores.\cite{OSF2018} Extending this approach to other conditions where motion is predictive of survival is only constrained by the availability of suitable training cases with known outcomes.      

\section*{Methods}
\subsection*{Study Population}
In a single-centre observational study, we analysed data collected from patients referred to the National Pulmonary Hypertension Service at the Imperial College Healthcare NHS Trust between May 2004 and October 2017. The study was approved by the Heath Research Authority and all participants gave written informed consent. Criteria for inclusion were a documented diagnosis of Group 4 PH investigated by right heart catheterization (RHC) and non-invasive imaging. All patients were treated in accordance with current guidelines including medical and surgical therapy as clinically indicated.\cite{Galie2016}

\begin{figure}[ht]
\centering
\includegraphics[width=\linewidth]{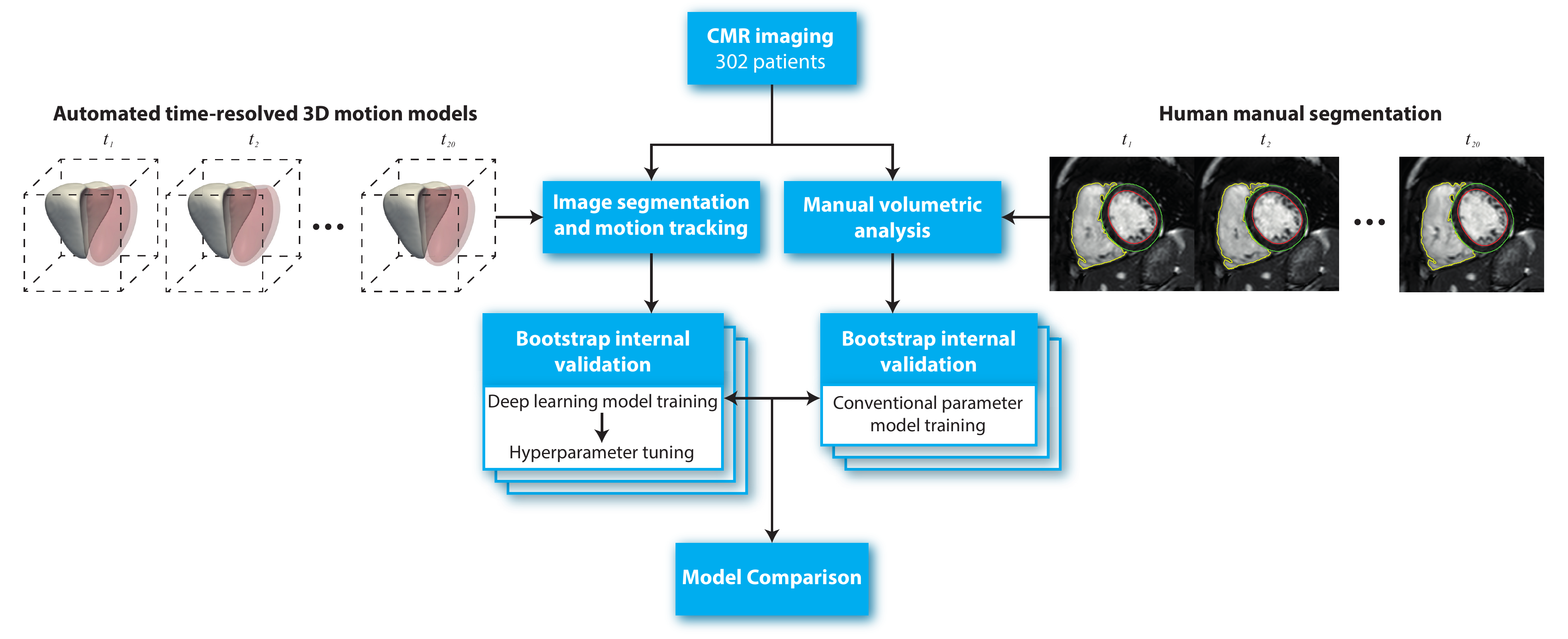}
\caption{Flowchart to show the design of the study. In total 302 patients with CMR imaging had both manual volumetric analysis and automated image segmentation (right ventricle shown in solid white, left ventricle in red) across 20 temporal phases ($t=1,..,20$). Internal validity of the predictive performance of a conventional parameter model and a deep learning motion model was assessed using bootstrapping. CMR, cardiac magnetic resonance.}
\label{fig:flowchart}
\end{figure}

\subsection*{MR Image Acquisition, Processing and Computational Image Analysis}
The CMR protocol has been previously described in detail.\cite{Dawes2017} Briefly, imaging was performed on a 1.5 T Achieva (Philips, Best, Netherlands), using a standard clinical protocol based on international guidelines.\cite{Kramer2015} The specific images analysed in this study were retrospectively-gated cine sequences, in the short axis plane of the heart, with a reconstructed spatial resolution of 1.3 x 1.3 x 10.0 mm and a typical temporal resolution of 29 ms. Images were stored on an open source data management system.\cite{woodbridge2013} Manual volumetric analysis of the images was independently performed by accredited physicians using proprietary software (cmr42, Circle Cardiovascular Imaging, Calgary, Canada) according to international guidelines with access to all available images for each subject and no analysis time constraint.\cite{Schulz2013} The derived parameters included the strongest and most well-established CMR findings for prognostication reported in a disease-specific meta-analysis.\cite{baggen2016cardiac}

We developed a CNN combined with image registration for shape-based biventricular segmentation of the CMR images. The pipeline method has three main components: segmentation, landmark localisation and shape registration. Firstly, a 2.5D multi-task FCN is trained to effectively and simultaneously learn segmentation maps and landmark locations from manually labelled volumetric CMR images. Secondly, multiple high-resolution 3D atlas shapes are propagated onto the network segmentation to form a smooth segmentation model. This step effectively induces a hard anatomical shape constraint and is fully automatic due to the use of predicted landmarks from the network.

We treat the problem of predicting segmentations and landmark locations as a multi-task classification problem. First, let us formulate the learning problem as follows: we denote the input training dataset by $S=\{(U_i, R_i, L_i), i=1,...,N_t\}$, where $N_t$ is the sample size of the training data, $U_i=\{u^i_j,j=1,...,|U_i|\}$ is the raw input CMR volume, $R_i=\{r^i_j,j=1,...,|R_i|\}$, $r^i_j \in \{1,...,N_r\}$ are the ground truth region labels for volume $U_i$ ($N_r=5$ representing 4 regions and background), and $L_i=\{l^i_j,j=1,...,|L_i|\}$, $l^i_j \in \{1,...,N_l\}$ are the labels representing ground truth landmark locations for $U_i$ ($N_l=7$ representing 6 landmark locations and background). Note that $|U_i|=|R_i|=|L_i|$ stands for the total number of voxels in a CMR volume. Let $\textbf{W}$ denote the set of all network layer parameters. In a supervised setting, we minimise the following objective function via standard (backpropagation) stochastic gradient descent (SGD):

\begin{equation} \label{eq:SDLloss}
L(\textbf{W}) = L_S(\textbf{W}) + \alpha L_D(\textbf{W}) + \beta L_L(\textbf{W}) + \gamma\| \textbf{W}\|_F^2,
\end{equation}
where $\alpha$, $\beta$ and $\gamma$ are weight coefficients balancing the four terms. $L_S(\textbf{W})$ and $L_D(\textbf{W})$ are the region-associated losses that enable the network to predict segmentation maps. $L_L(\textbf{W})$ is the landmark-associated loss for predicting landmark locations. $\|\textbf{W}\|_F^2$, known as the weight decay term, represents the Frobenius norm on the weights $\textbf{W}$. This term is used to prevent the network from overfitting. The training problem is therefore to estimate the parameters $\textbf{W}$ associated with all the convolutional layers. By minimising (\ref{eq:SDLloss}), the network is able to simultaneously predict segmentation maps and landmark locations. The definitions of the loss functions $L_S(\textbf{W})$, $L_D(\textbf{W})$ and $L_L(\textbf{W})$, used for predicting landmarks and segmentation labels, have been described previously.\cite{duan3D2018}

The FCN segmentations are  used to perform a non-rigid registration using cardiac atlases built from $>$1000 high resolution images,\cite{Bai2015} allowing shape constraints to be inferred. This approach produces accurate, high-resolution and anatomically smooth segmentation results from input images with low through-slice resolution  thus preserving clinically-important global anatomical features.\cite{duan3D2018} Motion tracking was performed for each subject using a 4D spatio-temporal B-spline image registration method with a sparseness regularisation term.\cite{shi2013temporal} The motion field estimate is represented by a displacement vector at each voxel and at each time frame $t=1,..,20$. Temporal normalisation was performed before motion estimation to ensure consistency across the cardiac cycle.

\begin{figure}[ht]
\centering
\includegraphics[width=\linewidth]{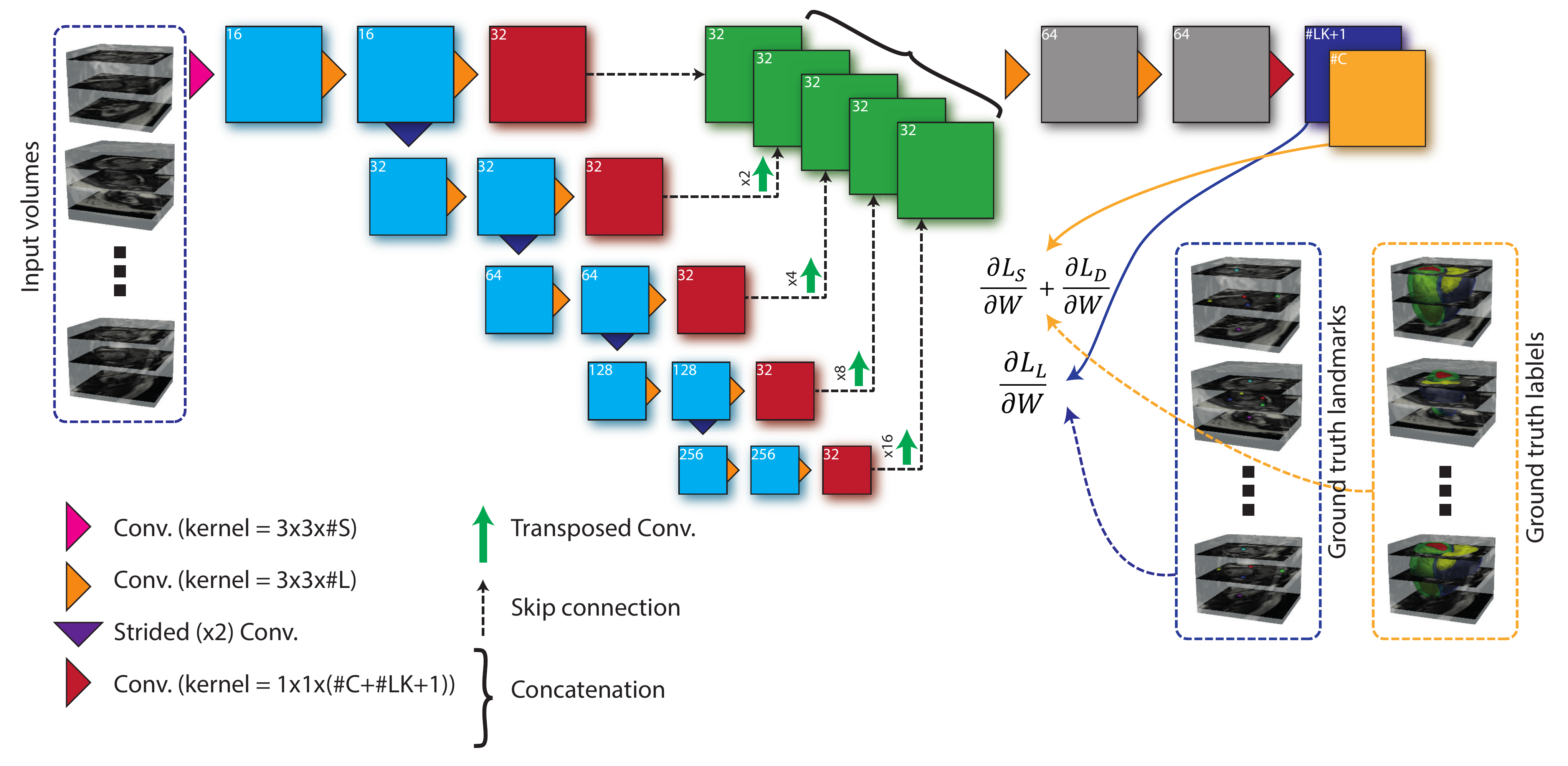}
\caption{The architecture of the segmentation algorithm. A fully convolutional network takes each stack of cine images as an input, applies a branch of convolutions, learns image features from fine to coarse levels, concatenates multi-scale features and finally predicts the segmentation and landmark location probability maps simultaneously. These maps, together with the ground truth landmark locations and label maps, are then used in the loss function (see Equation \ref{eq:SDLloss}) which is minimised via stochastic gradient descent.}
\label{fig:segpipeline}
\end{figure}

Spatial normalisation of each patient's data was achieved by registering the motion fields to a template space.  A template image was built by registering the high-resolution atlases at the end-diastolic frame and then computing an average intensity image. In addition, the corresponding ground-truth segmentations for these high-resolution  images were averaged to form a segmentation of the template image. A template surface mesh was then reconstructed from its segmentation using a 3D surface reconstruction algorithm. The motion field estimate lies within the reference space of each subject and so to enable inter-subject comparison all the segmentations were aligned to this template space by non-rigid B-spline image registration.\cite{rueckert1999nonrigid} We then warped the template mesh using the resulting non-rigid deformation and mapped it back to the template space. Twenty surface meshes, one for each temporal frame, were subsequently generated by applying the estimated motion fields to the warped template mesh accordingly. Consequently, the surface mesh of each subject at each frame contained the same number of vertices ($18,028$) which maintained their anatomical correspondence across temporal frames, and across subjects (Figure \ref{fig:segpipeline}). \\

\subsection*{Characterization of right ventricular motion}
The time-resolved 3D meshes described in the previous section were used to produce a relevant representation of cardiac motion - in this example of right-side heart failure limited to the RV. For this purpose, we utilized a sparser version of the meshes (down-sampled by a factor of \textasciitilde90) with 202 vertices. Anatomical correspondence was preserved in this process by utilizing the same vertices across all meshes. To characterize motion, we adapted an approach outlined in Bai \textit{et al} (2015).\cite{Bai2015_2}

This approach is used to produce a simple numerical representation of the trajectory of each vertex, i.e. the path each vertex traces through space during a cardiac cycle (see Figure \ref{fig:surfacemesh_mag}B). Let $(x_{vt}, y_{vt}, z_{vt})$ represent the Cartesian coordinates of vertex $v$ ($v=1,..,202$) at the $t^{th}$ time frame ($t=1,..,20$) of the cardiac cycle. At each time frame $t=2,3,...,20$, we compute the coordinate-wise displacement of each vertex from its position at time frame 1. This yields the following one-dimensional input vector:

\begin{equation}
\bm{x} = \bigg( x_{vt}-x_{v1}, \quad y_{vt}-y_{v1}, \quad z_{vt}-z_{v1} \bigg)_{2 \le t \le 20}^{1 \le v \le 202}
\end{equation}
\pagebreak[0]
Vector $\bm{x}$ has length 11,514 ($3 \times 19 \times 202$), and was used as the input feature for our prediction network.

\begin{figure}[!hb]
\centering
\includegraphics[width=\linewidth]{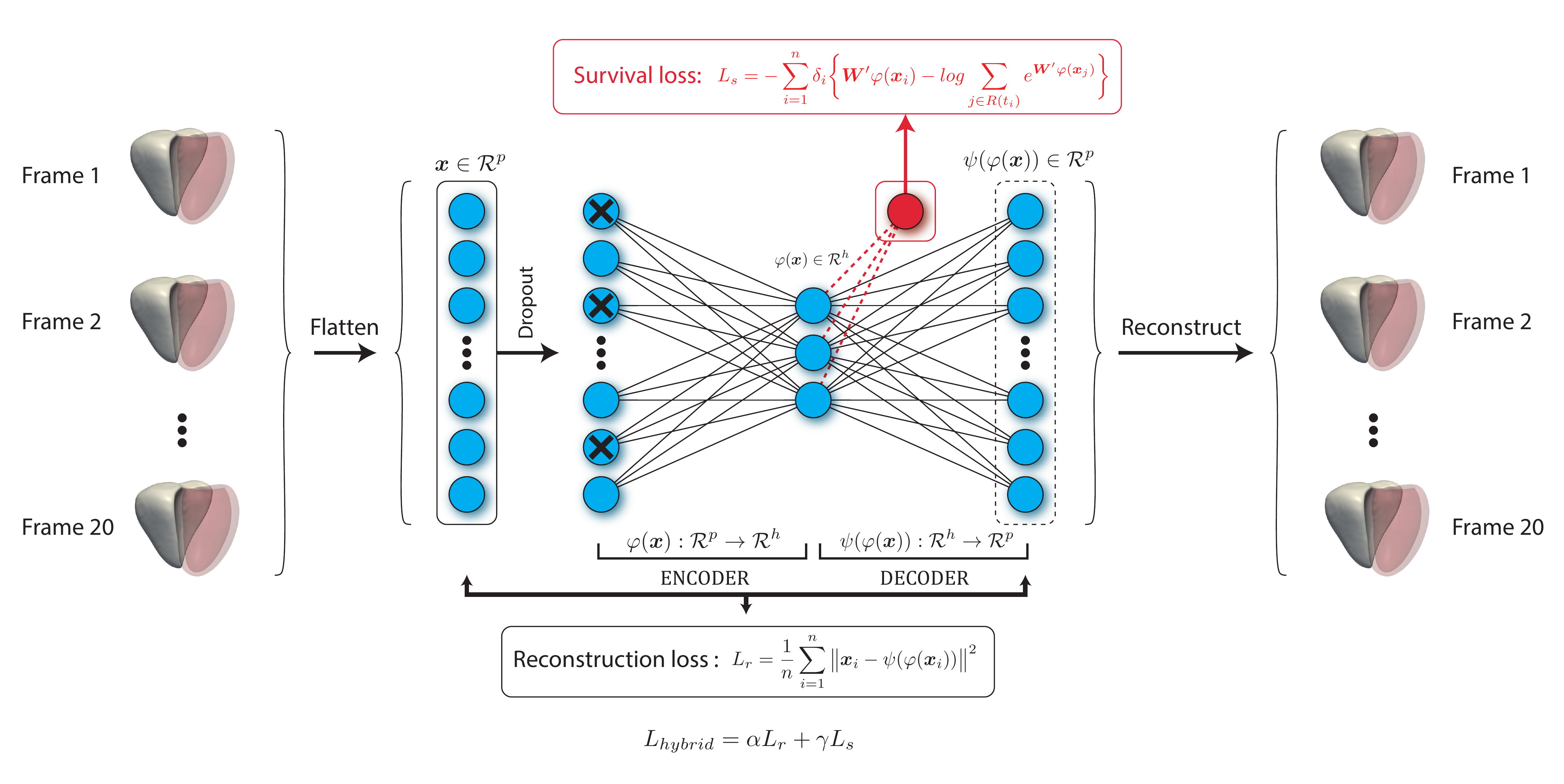}
\caption{The prediction network (4Dsurvival) is a denoising autoencoder that takes time-resolved cardiac motion meshes as its input (right ventricle shown in solid white, left ventricle in red). For the sake of simplicity two hidden layers, one immediately preceding and the other immediately following the central layer (latent code layer), have been excluded from the diagram. The autoencoder learns a task-specific latent code representation trained on observed outcome data, yielding a latent representation optimised for survival prediction that is robust to noise. The actual number of latent factors is treated as an optimisable parameter.}
\label{fig:autoenc}
\end{figure}

\subsection*{Network Design and Training}
Our 4D\textit{survival} network structure is summarized in Figure \ref{fig:autoenc}. We aimed to produce an architecture capable of learning a low-dimensional representation of RV motion that robustly captures prognostic features indicative of poor survival. The architecture's hybrid design combines a denoising autoencoder,\cite{Vincent2010} with a Cox proportional hazards model (described below).\cite{Cox1972} 

As before, we denote our input vector by $\bm{x} \in \mathbb{R}^{d_p}$, where $d_p =$ 11,514, the input dimensionality. Our network is based on a DAE, an autoencoder variant which learns features robust to noise.\cite{Vincent2010} The input vector $\bm{x}$ feeds directly into the encoder, the first layer of which is a stochastic masking filter that produces a corrupted version of $\bm{x}$. The masking is implemented using random dropout,\cite{Srivastava2014} i.e. we randomly set a fraction $m$ of the elements of vector $\bm{x}$ to zero (the value of $m$ is treated as an optimizable network parameter). The corrupted input from the masking filter is then fed into a hidden layer, the output of which is in turn fed into a central layer. This central layer represents the latent code, i.e. the encoded/compressed representation of the input. This central layer is referred to as the `code', or `bottleneck' layer. Therefore we may consider the encoder as a function $\phi(\cdot)$ mapping the input $\bm{x} \in \mathbb{R}^{d_p}$ to a latent code $\phi(\bm{x}) \in \mathbb{R}^{d_h}$, where $d_h \ll d_p$ (for notational convenience we consider the corruption step as part of the encoder). This produces a compressed representation whose dimensionality is much lower than that of the input (an undercomplete representation).\cite{Goodfellow2016} Note that the number of units in the encoder's hidden layer, and the dimensionality of the latent code ($d_h$) are not predetermined but, rather, treated as optimisable network parameters. The latent code $\phi(\bm{x})$ is then fed into the second component of the DAE, a multilayer decoder network that upsamples the code back to the original input dimension $d_p$. Like the encoder, the decoder has one intermediate hidden layer that feeds into the final layer, which in turn outputs a decoded representation (with dimension $d_p$ matching that of the input). The size of the decoder's intermediate hidden layer is constrained to match that of the encoder network, to give the autoencoder a symmetric architecture. Dissimilarity between the original (uncorrupted) input $\bm{x}$ and the decoder's reconstructed version (denoted here by $\psi(\phi(\bm{x}))$) is penalized by minimizing a loss function of general form $L(\bm{x}, \psi(\phi(\bm{x})))$. Herein, we chose a simple mean squared error form for $L$:

\begin{equation} \label{eq:reconloss}
L_{r} = \frac{1}{n}\sum_{i=1}^{n}\Vert\bm{x}_i - \psi(\phi(\bm{x}_i)) \Vert^2,
\end{equation}

where $n$ represents the sample size. Minimizing this loss forces the autoencoder to reconstruct the input from a corrupted/incomplete version, thereby facilitating the generation of a latent representation with robust features. Further, to ensure that these learned features are actually relevant for survival prediction, we augmented the autoencoder network by adding a prediction branch. The latent representation learned by the encoder $\phi(\bm{x})$ is therefore linked to a linear predictor of survival (see equation \ref{eq:coxmodel} below), in addition to the decoder. This encourages the latent representation $\phi(\bm{x})$ to contain features which are simultaneously robust to noisy input and salient for survival prediction. The prediction branch of the network is trained with observed outcome data, i.e. survival/follow-up time. For each subject, this is time elapsed from MRI acquisition until death (all-cause mortality), or if the subject is still alive, the last date of follow-up. Also, patients receiving surgical interventions were censored at the date of surgery. This type of outcome is called a \textit{right-censored time-to-event} outcome,\cite{Faraggi1995} and is typically handled using survival analysis techniques, the most popular of which is Cox's proportional hazards regression model:\cite{Cox1972}
\pagebreak[0]
\begin{equation} \label{eq:coxmodel}
log \frac{h_i(t)}{h_0(t)} = {\beta}_1 z_{i1} + {\beta}_2 z_{i2} + .... + {\beta}_p z_{ip}
\end{equation}

\noindent Here, $h_i(t)$ represents the hazard function for subject $i$, i.e the `chance' (normalized probability) of subject $i$ dying at time $t$. The term $h_0(t)$ is a baseline hazard level to which all subject-specific hazards $h_i(t)$ ($i=1,...,n$) are compared. The key assumption of the Cox survival model is that the \textit{hazard ratio} $h_i(t)/h_0(t)$ is constant with respect to time (\textit{proportional hazards} assumption).\cite{Cox1972} The natural logarithm of this ratio is modeled as a weighted sum of a number of predictor variables (denoted here by $z_{i1},...,z_{ip}$),  where the weights/coefficients are unknown parameters denoted by ${\beta}_1,...,{\beta}_p$. These parameters are estimated via maximization of the Cox proportional hazards partial likelihood function:

\begin{equation} \label{eq:coxpl}
log \mathcal{L}(\bm{\beta}) = \sum_{i=1}^{n} \delta_{i} \bigg\{\bm{\beta}'\bm{z}_i  - log \sum_{j \in R(t_i)} e^{\bm{\beta}' \bm{z}_j} \bigg\}
\end{equation}

\noindent In the expression above, $\bm{z}_i$ is the vector of predictor/explanatory variables for subject $i$, $\delta_{i}$ is an indicator of subject $i$'s status ($0$=Alive, $1$=Dead) and $R(t_i)$ represents subject $i$'s risk set, i.e. subjects still alive (and thus at risk) at the time subject $i$ died or became censored ($\{j : t_j > t_i\}$).

We adapt this loss function for our neural network architecture as follows: 

\begin{equation} \label{eq:coxplDL}
L_{s} =  -\sum_{i=1}^{n} \delta_{i} \bigg\{\bm{W}'\phi(\bm{x}_i)  - log \sum_{j \in R(t_i)} e^{\bm{W}'\phi(\bm{x}_j)} \bigg\}
\end{equation}

The term $\bm{W}'$ denotes a (1 $\times$ $d_h$) vector of weights, which when multiplied by the $d_h$-dimensional latent code $\phi(\bm{x})$ yields a single scalar ($\bm{W}'\phi(\bm{x}_i)$) representing the survival prediction (specifically, natural logarithm of the hazard ratio) for subject $i$.  Note that this makes the prediction branch of our 4D\textit{survival} network essentially a simple linear Cox proportional hazards model, and the predicted output may be seen as an estimate of the log hazard ratio (see Equation \ref{eq:coxmodel}).
\\ 
\\
For our \textit{} network, we combine this survival loss with the reconstruction loss from equation \ref{eq:reconloss} to form a hybrid loss given by:

\begin{equation} \label{eq:hybridloss}
L_{hybrid} = \alpha L_{r} + \gamma L_{s} = \alpha \bigg[ \frac{1}{n}\sum_{i=1}^{n}\Vert\bm{x}_i - \psi(\phi(\bm{x}_i)) \Vert^2 \bigg]+      \gamma \bigg[ -\sum_{i=1}^{n} \delta_{i} \bigg\{\bm{W}'\phi(\bm{x}_i)  - log \sum_{j \in R(t_i)} e^{\bm{W}'\phi(\bm{x}_j)} \bigg\} \bigg]
\end{equation}
\\
The terms $\alpha$ and $\gamma$ are used to calibrate the contributions of each term to the overall loss, i.e. to control the tradeoff between survival prediction versus accurate input reconstruction. During network training, they are treated as optimisable network hyperparameters, with $\gamma$ chosen to equal $1 - \alpha$ for convenience. 

The loss function was minimized via backpropagation. To avoid overfitting and to encourage sparsity in the encoded representation, we applied $L_1$ regularization. The rectified linear unit (ReLU) activation function was used for all layers, except the prediction output layer (linear activation was used for this layer). Using the adaptive moment estimation (\textit{Adam}) algorithm, the network was trained for 100 epochs with a batch size of 16 subjects. The learning rate is treated as a hyperparameter (see Table \ref{tab:hyperp}). During training, the random dropout (input corruption) was repeated at every backpropagation pass. The network was implemented and trained in the Python deep learning libraries \textit{TensorFlow} \cite{Abadi2015} and \textit{Keras} \cite{Chollet2015}, on a high-performance computing cluster with an Intel Xeon E5-1660 CPU and NVIDIA TITAN Xp GPU. The entire training process (including hyperparameter search and bootstrap-based internal validation [see subsections below]) took a total of 76 hours.

\subsection*{Hyperparameter Tuning}
To determine optimal hyperparameter values, we utilized particle swarm optimization (PSO),\cite{Kennedy1995} a gradient-free meta-heuristic approach to finding optima of a given objective function. Inspired by the social foraging behavior of birds, PSO is based on the principle of \textit{swarm intelligence}, which refers to problem-solving ability that arises from the interactions of simple information-processing units.\cite{Engelbrecht2005} In the context of hyperparameter tuning, it can be used to maximize the prediction accuracy of a model with respect to a set of potential hyperparameters.\cite{Lorenzo2017} We used PSO to choose the optimal set of hyperparameters from among predefined ranges of values (summarized in Table \ref{tab:hyperp}). We ran the PSO algorithm for 50 iterations, at each step evaluating candidate hyperparameter configurations using 6-fold cross-validation. The hyperparameters at the final iteration were chosen as the optimal set. This procedure was implemented via the Python library \textit{Optunity}.\cite{Claesen2014} 

\begin{table}[!b]
\centering
\begin{tabular}{p{5.5cm} | c}
\textbf{Hyperparameter} & \textbf{Search Range} \\
\hline
\hspace{3mm}Dropout & [0.1, 0.9] \\
\hspace{3mm}\# of nodes in hidden layers & [75, 250] \\
\hspace{3mm}Latent code dimensionality ($h$) & [5, 20] \\
\hspace{3mm}Reconstruction loss penalty ($\alpha$) & [0.3, 0.7] \\
\hspace{3mm}Learning Rate & [$10^{-6}$, $10^{-4.5}$] \\
\hspace{3mm}$L_1$ regularization penalty & [$10^{-7}$, $10^{-4}$] \\
\hline
\end{tabular}
\caption{\label{tab:hyperp}Hyperparameter search ranges}
\end{table}

\subsection*{Model Validation and Comparison}
\medskip
\subsubsection*{Predictive Accuracy Metric}
Discrimination was evaluated using Harrell's concordance index,\cite{Harrell1982} an extension of area under the receiver operating characteristic curve (AUC) to censored time-to-event data:

\begin{equation} \label{eq:cindex}
C = \frac{\sum_{i,j} {\delta}_i \cdot I({\eta}_i > {\eta}_j) \cdot I(t_i < t_j) }{\sum_{i,j} {\delta}_i \cdot I(t_i < t_j)}
\end{equation}

In the above equation, the indices $i$ and $j$ refer to pairs of subjects in the sample and $I()$ denotes an indicator function that evaluates to 1 if its argument is true (and 0 otherwise). Symbols ${\eta}_i$ and ${\eta}_j$ denote the predicted risks for subjects $i$ and $j$. The numerator tallies the number of subject pairs $(i,j)$ where the pair member with greater predicted risk has shorter survival, representing agreement (concordance) between the model's risk predictions and ground-truth survival outcomes. Multiplication by ${\delta}_i$ restricts the sum to subject pairs where it is possible to determine who died first (i.e. informative pairs). The $C$ index therefore represents the fraction of informative pairs exhibiting concordance between predictions and outcomes. In this sense, the index has a similar interpretation to the AUC (and consequently, the same range).

\subsubsection*{Internal Validation}
In order to get a sense of how well our model would generalize to an external validation cohort, we assessed its predictive accuracy within the training sample using a bootstrap-based procedure recommended in the guidelines for \textit{Transparent Reporting of a multivariable model for Individual Prognosis Or Diagnosis} (TRIPOD).\cite{Moons2015} This procedure attempts to derive realistic, `optimism-adjusted' estimates of the model's generalization accuracy using the training sample.\cite{Harrell1996} Below, we outline the steps of the procedure:

\begin{description}[labelindent=0.5cm, leftmargin=1.8cm]

\item[(\textbf{Step 1})]A prediction model was developed on the full training sample (size $n$), utilizing the hyperparameter search procedure discussed above to determine the best set of hyperparameters. Using the optimal hyperparameters, a final model was trained on the full sample. Then the Harrell’s concordance index ($C$) of this model was computed on the full sample, yielding the \textit{apparent} accuracy, i.e. the inflated accuracy obtained when a model is tested on the same sample on which it was trained/optimized.

\item[(\textbf{Step 2})]A bootstrap sample was generated by carrying out $n$ random selections (with replacement) from the full sample. On this bootstrap sample, we developed a model (applying exactly the same training and hyperparameter search procedure used in Step 1) and computed $C$ for the bootstrap sample (henceforth referred to as \textit{bootstrap performance}). Then the performance of this bootstrap-derived model on the original data (the full training sample) was also computed (henceforth referred to as \textit{test performance})

\item[(\textbf{Step 3})]For each bootstrap sample, the optimism was computed as the difference between the bootstrap performance and the test performance.

\item[(\textbf{Step 4})]Steps 2-3 were repeated $B$ times (where $B$=50).

\item[(\textbf{Step 5})]The optimism estimates derived from Steps 2-4 were averaged across the $B$ bootstrap samples and the resulting quantity was subtracted from the apparent predictive accuracy from Step 1.
\end{description}

This procedure yields an optimism-corrected estimate of the model's concordance index:

\begin{equation}\label{eq:optcorrC}
C_{corrected} = C_{full}^{full} - \frac{1}{B}\sum_{b=1}^{B} \bigg( C_{b}^{b} - C_{b}^{full} \bigg)
\end{equation}

Above, symbol $C_{s_{1}}^{s_{2}}$ refers to the concordance index of a model trained on sample $s_1$ and tested on sample $s_2$. The first term refers to the \textit{apparent} predictive accuracy, i.e. the (inflated) concordance index obtained when a model trained on the full sample is then tested on the same sample. The second term is the average \textit{optimism} (difference between \textit{bootstrap performance} and \textit{test performance}) over the $B$ bootstrap samples. It has been demonstrated that this sample-based average is a nearly unbiased estimate of the expected value of the optimism that would be observed in external validation.\cite{Efron1983,Efron1993,Harrell1996,Smith2014} Subtraction of this optimism estimate from the apparent predictive accuracy gives the optimism-corrected predictive accuracy.

\subsubsection*{Conventional Parameter model}
As a benchmark comparison to our RV motion model, we trained a Cox proportional hazards model using conventional RV volumetric indices including RV end-diastolic volume (RVEDV), RV end-systolic volume (RVESV) and the difference between these measures expressed as a percentage of RVEDV, RV ejection fraction (RVEF) as survival predictors. To account for collinearity among these predictor variables, an $L_2$-norm regularization term was added to the Cox partial likelihood function:

\begin{equation} \label{eq:coxl2}
log L(\beta) = \sum_{i=1}^{n} \delta_{i} \bigg\{\bm{\beta}'\bm{x}_i  - log \sum_{j \in R(t_i)} e^{\bm{\beta}' \bm{x}_j} \bigg\} + \frac{1}{2}{\lambda}\Vert{\bm{\beta}}{\Vert}^2
\end{equation}

In the equation above, $\lambda$ is a parameter that controls the strength of the penalty. The optimal value of $\lambda$ was selected via cross-validation. The Cox model was set up and fit using the Python library \textit{lifelines}.\cite{lifelines2018}

\subsubsection*{Model Interpretation}
To facilitate interpretation of our 4D\textit{survival} network we used Laplacian Eigenmaps to project the learned latent code into two dimensions,\cite{Belkin2002} allowing latent space visualization. Neural networks derive predictions through multiple layers of nonlinear transformations on the input data. This complex architecture does not lend itself to straightforward assessment of the relative importance of individual input features. To tackle this problem we used a simple regression-based inferential mechanism to evaluate the contribution of motion in various regions of the RV to the model's predicted risk. For each of the 202 vertices in our RV mesh models we computed a single summary measure of motion by averaging the displacement magnitudes across 19 frames. This yielded one mean displacement value per vertex. This process was repeated across all subjects. Then we regressed the predicted risk scores onto these vertex-wise mean displacement magnitude measures using a mass univariate approach, i.e. for each vertex \textit{v}  ($v = 1,...,202$), we fitted a linear regression model where the dependent variable was predicted risk score, and the independent variable was average displacement magnitude of vertex \textit{v}. Each of these 202 univariate regression models was fitted on all subjects and yielded one regression coefficient representing the effect of motion at a vertex on predicted risk. The absolute values of these coefficients, across all vertices, were then mapped onto a template RV mesh to provide a visualization of the differential contribution of various anatomical regions to predicted risk.

\section*{Data and code availability}
Algorithms, motion models and statistical analysis are publicly available under a GNU General Public License.\cite{DeepSurvival} A training simulation is available as a \textit{Docker} image with an interactive \textit{Jupyter} notebook hosted on \textit{Binder}. Personal data are not available due to privacy restrictions.

\bibliography{NatMachIntell3}

\section*{Acknowledgements}

The research was supported by the British Heart Foundation (NH/17/1/32725, RE/13/4/30184); National Institute for Health Research (NIHR) Biomedical Research Centre based at Imperial College Healthcare NHS Trust and Imperial College London; and the Medical Research Council, UK. The TITAN Xp GPU used for this research was kindly donated by the NVIDIA Corporation.

\section*{Author contributions statement (CRediT)}

G.A.B., C.B. and T.J.W.D. - methodology, software, formal analysis and writing original draft. J.D. - methodology and software; A. de M. - formal analysis; L.S.G.E.H., J.S.R.G, M.R.W and S.A.C. - investigation; D.R. - software and supervision; D.P.O'R. - conceptualization, supervision, writing – review and editing, funding acquisition. All authors reviewed the final manuscript. 

\section*{Additional information}

The authors declare no competing financial interests.

\end{document}